\documentclass[runningheads]{llncs}

\usepackage{eccv}

\usepackage{eccvabbrv}

\usepackage{graphicx}
\usepackage{booktabs}

\usepackage[accsupp]{axessibility}  %

\usepackage{hyperref}

\usepackage{orcidlink}

\usepackage{multirow}
\usepackage{multicol}

\usepackage{colortbl}
\usepackage{xcolor} 
\usepackage{makecell}

\usepackage{algorithm}
\usepackage{algpseudocode}
\usepackage{amsmath}
\usepackage{amssymb}

\usepackage{wrapfig}

\begin{document}

\title{Mapping the Concept Landscape: Structural Perception of Global Distributions for Transparent Data Pruning} 

\titlerunning{Structural Perception of Global Distributions for Transparent Data Pruning}

\author{Dongyue Wu\inst{1,3}\textsuperscript{*}\orcidlink{0000-0002-4716-856X} \and
Tao Ma\inst{1,2}\textsuperscript{*}\orcidlink{0009-0002-3995-3207}}
\authorrunning{D.~Wu et al.}

\institute{State Key Laboratory of Multispectral Information Intelligent Processing Technology, School of Artificial Intelligence and Automation,\\ Huazhong University of Science and Technology, Wuhan, China\\
\and 
Institute of Computing Technology, Chinese Academy of Sciences, Beijing, China
\and
Ant Group \\
\email{\{dongyue\_wu, u202215201\}@hust.edu.cn}}

\maketitle
{
\def\thefootnote{*} %
\footnotetext{Equal contribution.} 
}

\begin{abstract}
Existing data pruning methods predominantly rely on high-dimensional feature embeddings to measure sample importance. However, these compressed vectors often obscure fine-grained semantic interactions, leading to suboptimal coverage of rare semantic concepts in the pruned subsets. In this paper, we propose Mapping the Concept Landscape (MCL), a novel structural perception framework for transparent data pruning. Instead of abstract embeddings, we represent each image-caption pair as an explicit sample-level graph comprising entities, events, and attributes. By integrating these individual graphs into a comprehensive dataset-level graph, we characterize the global distribution of semantic concepts and quantify their rarity across the entire corpus. Based on this structured perception, we develop a greedy concept-coverage maximization algorithm that iteratively selects samples to maximize the marginal gain of high-value, under-represented concepts. Experimental results on various benchmarks demonstrate that our method not only achieves superior pruning efficiency compared to state-of-the-art methods but also provides a transparent and interpretable audit trail for the selection process. 
\end{abstract}

\section{Introduction}

The rapid expansion of web-scale vision–language datasets, such as LAION \cite{schuhmann2022laion} and CC12M \cite{changpinyo2021conceptual}, has been a key driver behind the recent success of large vision-language models \cite{dai2023instructblip,radford2021learning}. However, this growth also introduces practical challenges. Training on massive datasets incurs prohibitive storage and computational costs, while the automated data collection process inevitably introduces a large amount of redundant, low-quality, and noisy samples. As a result, data pruning~\cite{dataset_pruning, EL2N}, has become a crucial technique for improving data efficiency, aiming to select a compact yet representative subset that preserves model performance under strict data budgets.

At its core, dataset pruning is an optimization problem: given a limited budget, which subset of samples should be retained to best preserve the information of the original dataset? Most existing pruning methods address this problem at the level of sample-wise representations, selecting data based on loss value~\cite{rho,infobatch,EL2N}, uncertainty~\cite{svp,dynunc,PBDP,raju}, learning difficulty~\cite{fogetting,sorscher2022beyond}, or influence~\cite{influence} measured in a compressed sample-level embedding space. Other works~\cite{datatailor,coincide,DivBS,kcenter,pfb} attempt to preserve dataset diversity by selecting samples according to similarity or distance between embeddings. These methods implicitly assume that geometric similarity in the embedding space faithfully reflects semantic redundancy in the underlying data.

\begin{figure}[t] 
\centering         
\includegraphics[width=\textwidth]{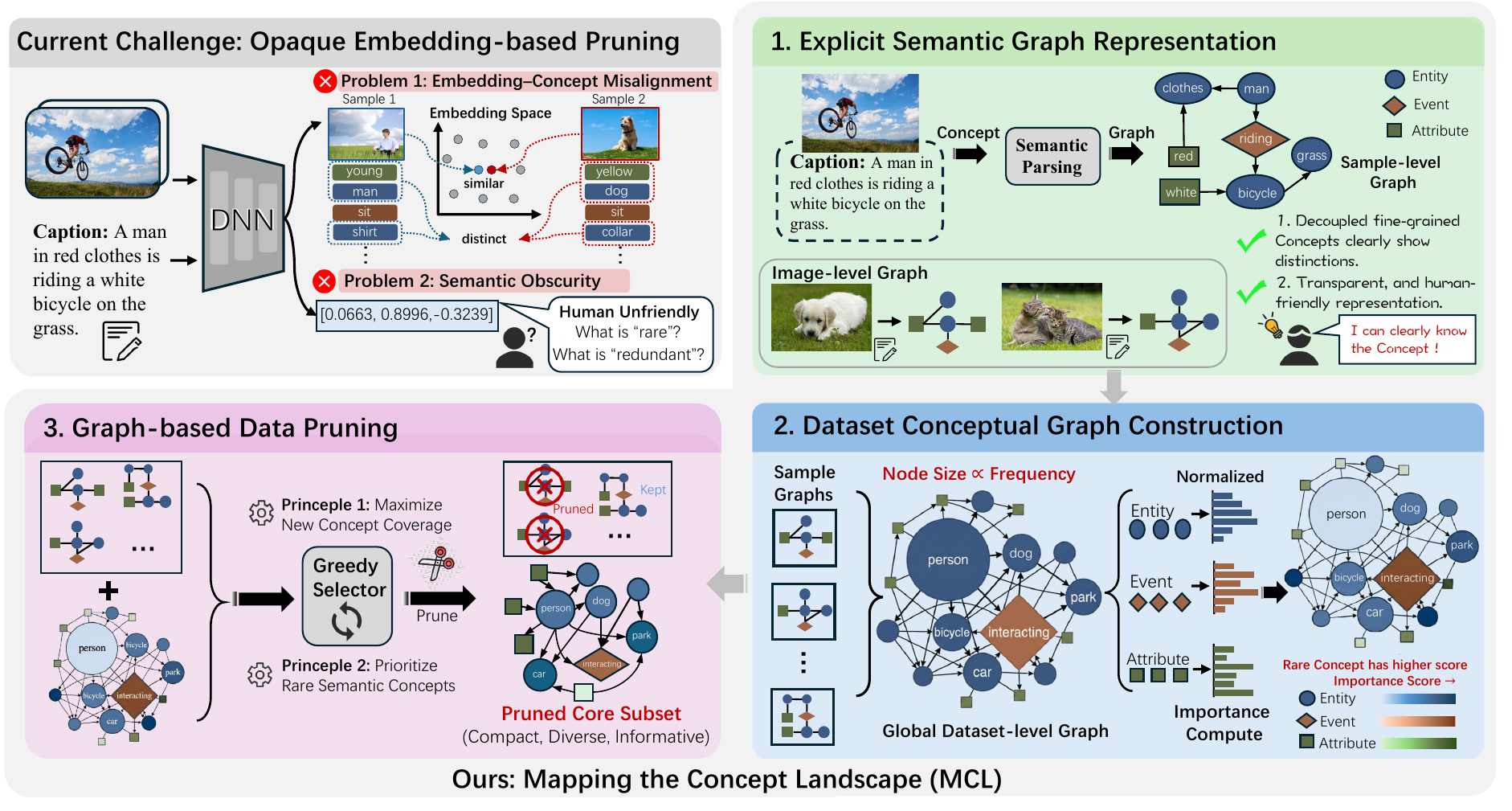}
\caption{Illustration of the challenge of data pruning methods and the overview of our proposed Mapping the Concept Landscape~(MCL) framework for transparent data pruning.
MCL aims to maximize semantic coverage based on concept graphs. The node size is positively related to the number of samples a concept appears in the dataset. } 
\label{fig:pipline}
\end{figure}

However, this assumption leads to a fundamental \textit{Embedding–Concept Misalignment}. While embedding similarity measures proximity between samples in a learned geometric space, semantic diversity is inherently compositional and structured at the concept level. Consequently, redundancy in the embedding space does not reliably correspond to redundancy in semantic concepts. Samples that appear redundant in representation space may still contain distinct fine-grained concepts whereas geometrically diverse samples may largely overlap in their conceptual content. This mismatch arises because rich multi-concept semantics are collapsed into a single coupled vector representation. Hence, embedding-based pruning methods primarily optimize geometric diversity rather than true semantic coverage, and may systematically discard low-fraction yet semantically important concepts even when the retained subset appears well-distributed in embedding space.

Moreover, compressing each image–text pair into a dense numerical embedding introduces another practical limitation: a lack of interpretability at the dataset level. Pruning decisions are made in an opaque embedding space, making it unclear for human which semantic concepts are preserved or discarded. While such representations may be effective for model optimization, they are difficult for humans to interpret. As a result, practitioners have limited ability to audit, inspect, and refine pruning outcomes or reason about the resulting semantic distribution of the whole dataset.

Together, these limitations stem from treating samples as indivisible units and obscuring their internal semantic structure, motivating a shift from sample-level pruning to concept-level coverage.
In this paper, we propose \textit{Mapping the Concept Landscape} (MCL), a graph-structured framework for multimodal dataset pruning that explicitly models and optimizes concept-level coverage. Instead of representing each sample as an opaque vector, MCL maps each image–text pair to an interpretable sample-level semantic graph composed of atomic concepts, including Entities, Events, and Attributes. While concept extraction may be imperfect at the individual sample level, MCL operates on aggregated concept statistics across the dataset, where relative frequencies and coverage patterns dominate pruning decisions. By aggregating sample-level graphs, MCL constructs a dataset-level concept graph that provides a transparent, human-friendly view of the dataset’s semantic landscape, enabling direct inspection of concept distributions, redundancies, and gaps.
Based on this dataset-level representation, pruning is formulated as a coverage maximization problem over semantic concepts. Concepts that appear frequently contribute diminishing additional value, whereas underrepresented concepts provide higher marginal benefit when introduced into the selected subset. This naturally leads to a greedy selection strategy that prioritizes samples according to their marginal contribution, namely the extent to which they introduce previously uncovered, high-value concepts. Importantly, emphasizing rare concepts in this manner does not imply over-sampling the long tail. Since concept coverage is a monotone and saturating objective, the marginal gain of repeatedly selecting samples containing the same concepts quickly diminishes, enabling the greedy procedure to balance redundancy reduction with the preservation of dominant semantics. 
In summary, our main contributions are three-fold:
\begin{itemize}
\item We propose MCL, a data pruning framework that moves beyond opaque embedding-based criteria by representing data using structured  concept graphs, enabling effective pruning and human-interpretable dataset analysis.
\item We introduce a greedy pruning algorithm that selects samples based on their marginal contribution to concept coverage, prioritizing underrepresented concepts and reducing redundant and dominant semantics.
\item Extensive experiments demonstrate that MCL achieves state-of-the-art pruning efficiency: using less than 10\% of training time and selected data, our method attains performance comparable to full-data baselines.
\end{itemize}

\section{Related Work}

\noindent \textbf{Dataset Pruning and Coreset Selection.}
Dataset pruning and coreset selection aim to identify a compact subset of training data that preserves the performance of the full dataset. Early studies explored this idea in contexts such as active learning\cite{bordes2005fast,sener2017active}, noisy learning\cite{mirzasoleiman2020coresets}, and continual learning\cite{rebuffi2017icarl,aljundi2019gradient}. Many approaches estimate sample importance using signals derived from training dynamics\cite{fogetting,EL2N,wu2023self}, loss values\cite{sb,rho,infobatch,EL2N,loshchilov2015online}, gradient\cite{EL2N,dataset_pruning,moso,DivBS}. Other works focus on clustering or the distance to centers~\cite{sorscher2022beyond,ccs} are used to select representative instances. These methods have demonstrated effectiveness in reducing dataset size while maintaining model performance. However, most existing techniques rely on compressed feature embeddings to define pruning metrics, treating each sample as an indivisible unit summarized by a single vector representation.

\smallskip
\noindent \textbf{Data Selection for Multimodal Data.}
With the emergence of large vision–language models, data selection has recently attracted attention in visual instruction tuning. Several methods~\cite{datatailor,wei2023instructiongpt,chen2024your,xia2024less,yang2024smalltolarge,zhou2023lima} attempt to evaluate the usefulness of multimodal instruction data through model-driven signals. For example, Self-Filter\cite{selffilter} assigns scores to instruction data using a learned scoring network, while TIVE\cite{tive} estimates sample importance through gradient-based analysis of the target model. Other approaches leverage auxiliary models~\cite{yang2024smalltolarge} or clustering techniques~\cite{coincide, datatailor} to identify important samples. Methods like ICONS~\cite{icons} explores the size allocation across different instruction tasks. 

\smallskip
\noindent \textbf{Diversity and Coverage-Based Data Selection.}
A line of research emphasizes maintaining diversity or coverage in the selected subset~\cite{pfb,datatailor,DivBS,kcenter,infomax}. These approaches typically perform clustering or similarity-based sampling to ensure broader coverage of the data distribution. However, diversity is usually measured in embedding space, where complex samples are represented by single compressed vectors. Such representations often obscure the internal semantic structure of image–text pairs and make it difficult to reason about fine-grained concept coverage. In contrast, our method models datasets through explicit semantic concepts,
enabling interpretable and fine-grained coverage-aware pruning.

\section{Method}

We present Mapping the Concept Landscape (MCL), a concept-centric framework for multimodal dataset pruning. Given a dataset of image–text pairs, MCL explicitly models the dataset’s semantic structure at the concept level and selects a representative subset by maximizing fine-grained semantic coverage under a predefined budget.
As illustrated in Fig.~\ref{fig:pipline}, MCL consists of the following key steps: First, each sample is mapped to a sample-level concept graph that captures its fine-grained semantic content using explicit and interpretable atomic concepts. Second, these sample-level graphs are aggregated to construct a dataset-level concept graph, which summarizes the global semantic distribution of the dataset. Third, based on concept frequencies, we define a concept importance measure and formulate dataset pruning as a concept coverage maximization problem. Finally, we adopt a greedy selection strategy that iteratively selects samples according to their marginal contribution to the overall concept coverage. This formulation enables MCL to reduce redundancy while preserving diverse and semantically informative concepts, and simultaneously provides a transparent interface for understanding and auditing the pruned dataset.

\subsection{Sample-level Concept Graph Construction}

A central design choice of MCL is to represent each sample using an explicit, transparent, and fine-grained semantic representation rather than a single coupled and obscured embedding. To this end, we construct an image-level concept graph $G_i=(V_i,E_i)$ for each image–text pair $x_i=(I_i, T_i)$. The caption $T_i$ can be either the original accompanying caption or a generated caption produced by an off-the-shelf vision–language model. Each node $v$ in the nodes set $V_i$ serves as the atomic semantic unit for subsequent dataset-level distribution modeling. A single node $v$ represents an atomic concept, and edges in set $E_i$ encode the concept co-occurrence and dependency relationships within the caption $T_i$. We also categorize each node to one of the three concept types:
\begin{itemize}
    \item \textbf{Entities}, corresponding to nouns that denote objects or visual subjects;
    \item \textbf{Events}, corresponding to verbs that describe actions or interactions;
    \item \textbf{Attributes}, corresponding to adjectives or adverbs that specify properties, states, or modifiers of entities and events.
\end{itemize}
These three categories jointly capture the majority of fine-grained semantics commonly present in image–text descriptions, including objects, actions, attributes, and their contextual relations. Please refer to Appendix for visualizations. Compared to representing a sample as a single embedding vector, this graph-based representation explicitly decomposes the sample into interpretable semantic components, enabling fine-grained reasoning at the concept level. 

The specific procedure used to extract concepts and construct the image-level graph is not a contribution of this work and can be implemented using standard linguistic analysis tools. 
For example, we can either employ scene graph generation methods~\cite{krishna2017visual} based on $I_i$ or simply extract the conceptual graph from $T_i$ using dependency parsing methods~\cite{dozat2017deep, dozat2017stanford, peng2017deep}.
We find that employing dependency parsing on these captions is not only computationally efficient but also effective at mapping a wide range of conceptual details.
Therefore, we utilize the spaCy~\cite{honnibal2020spacy}, a widely-adopted industrial-strength NLP library, to perform text-based dependency parsing on each $x_i$ based on the text caption $T_i$. The process identifies and filters out stop words, retaining only semantically meaningful tokens. Based on the dependency tree and Part-of-Speech (POS) tags produced by spaCy, we construct the sample-level concept graph $G_i$. 
The image-level graph serves as a structured container that exposes the sample’s internal semantic composition, which can later be aggregated and analyzed at the dataset level.

\subsection{Dataset-level Concept Graph Construction}

While image-level graphs capture fine-grained semantics for individual samples, our framework requires modeling the semantic coverage and concept distribution at the scale of the entire dataset. To enable such global analysis, we aggregate all image-level graphs into a unified dataset-level concept graph.

Formally, given a dataset $\mathcal{D}=\{x_i\}^N_{i=1}$, we construct a dataset-level graph $G_{\mathcal{D}}=(V_{\mathcal{D}}, E_{\mathcal{D}})$ by taking the union of all concept nodes $V_D = \bigcup_{i=1}^{N} V_i$.
Each node $v_j\in V_D$ is associated with a weight that counts the number of samples where the corresponding concept appears:
\begin{equation}
\label{eq:node_weight}
w(v) = \sum_{i=1}^{N} \mathbf{1}[v \in V_i].
\end{equation}
This weight reflects the empirical fraction of the concept in the dataset and serves as a proxy for its redundancy or rarity.
The resulting dataset-level concept graph $G_{\mathcal{D}}$ provides an explicit summary of the dataset’s semantic landscape. Concepts with large weights correspond to frequently occurring semantics, while low-weight nodes indicate rare and scarce concepts. Unlike embedding-based summaries, this representation of the whole dataset is directly interpretable by humans and enables straightforward inspection of concept distributions and semantic relations within the dataset. The constructed $G_{\mathcal{D}}$ is a summary structure which employs the graph to model the concept distribution, rather than an object to be pruned or conduct data selection. Please note, the actual objects to be pruned are still the individual samples $x_i$.

Crucially, MCL operates on aggregated concept statistics rather than individual sample annotations. Although concept extraction at the image level may be noisy or imperfect, such errors are amortized when aggregated across the dataset. Since pruning decisions are driven by relative concept frequencies and coverage patterns, the dataset-level concept graph provides a stable and robust foundation for defining concept importance and guiding data selection, which we describe in the following sections.

\subsection{Concept Importance and Coverage Objective}

Based on the dataset-level concept graph, we now formalize dataset pruning as a concept coverage maximization problem. We first define a concrete instantiation of concept node importance, and then introduce the resulting coverage objective.

\smallskip
\noindent \textbf{Node Importance with Structural Diversity Awareness}. 
We first define a concept importance function $\phi(v)$ for each concept node $v \in V_{\mathcal{D}}$. Intuitively, concepts that occur frequently tend to encode common and thus redundant semantics, whereas rare concepts are more likely to capture distinctive or underrepresented information.
While $w(v)$ captures how common a concept is, it does not account for how the concept participates in the overall semantic structure of the dataset. A concept that appears rarely but only in isolated contexts may be less informative than a rare concept that interacts with many other concepts through diverse relationships. To capture both aspects, we define concept importance by jointly considering semantic rarity and structural participation. Specifically, for each concept node $v$, we define its unnormalized node importance score $\tilde{\phi}(v)$ as follows
\begin{equation}
\tilde{\phi}(v) = \log \left( \frac{N_{type(v)}}{w(v)} \right) \cdot [1 + \log (d(v))],
\end{equation}
where $N_{type(v)}$ denotes the total number of concept nodes within the same semantic category~(Entity, Event, Attribute) as $v$, and $d(v)$ is the degree of $v$ in $G_{\mathcal{D}}$.
The first term is an inverse-fraction factor that assigns higher importance to insufficiently represented concepts, while the second term serves as a lightweight relational correction, promoting concepts that participate in richer semantic interactions. This design allows relational information to be incorporated without explicitly modeling edge importance, resulting in a simpler and more efficient formulation.

\smallskip
\noindent \textbf{Category-wise Normalization}.
Concept nodes in the dataset graph belong to three semantic categories: Entity, Event, and Attribute. These categories naturally differ in scale and fraction distribution. To prevent any single category from dominating the importance scores, we perform normalization independently within each category.
Formally, the final concept importance is defined as
\begin{equation}
\phi(v) = \frac{\tilde{\phi}(v) - \min_{u \in V_{type(v)}} \tilde{\phi}(u)}{\max_{u \in V_{type(v)}} \tilde{\phi}(u) - \min_{u \in V_{type(v)}} \tilde{\phi}(u) + \epsilon'},
\label{eq:importance_norm}
\end{equation}
where $V_{type(v)}$ denotes the set of nodes belonging to the same category as $v$, and $\epsilon$ is a small constant for numerical stability.
This normalization ensures that importance scores are comparable across different concept types, allowing rare events or attributes to contribute meaningfully alongside entity concepts.

\smallskip
\noindent \textbf{Concept Coverage Function}.
Given a subset of selected samples $\mathcal{S} \subseteq \mathcal{D}$, we define its covered concept node set as
\begin{equation}
V_{\mathcal{S}} = \bigcup_{x_i \in \mathcal{S}}^{N} V_i,
\end{equation}
where $V_i$ denotes the set of concept nodes associated with sample $x_i$. The overall concept coverage function of $\mathcal{S}$ is then defined as 
\begin{equation}
F(S) = \sum_{v \in V_{\mathcal{S}}} \phi(v).
\end{equation}
This formulation explicitly accounts for semantic redundancy: once a concept is covered by the selected subset, additional samples containing the same concept do not further increase the objective value. Therefore, samples that introduce uncovered and structurally informative concepts yield higher marginal gains. 

\smallskip
\noindent \textbf{Dataset Pruning Objective}.
With the above definitions, dataset pruning can be formulated as the following constrained optimization problem:
\begin{equation}
\max_{S \subseteq D} F(S) \quad \text{s.t.} \quad |\mathcal{S}| \leq b,
\label{eq:objective}
\end{equation}
where $b=(1-p)\cdot|\mathcal{D}|$ denotes the target number of the retained subset after data selection and $p$ denotes the pruning ratio. By optimizing concept coverage under this objective, MCL prioritizes samples that jointly preserve rare, diverse, and relationally informative semantics, while maintaining a transparent and interpretable pruning criterion grounded in the dataset’s concept structure.

\subsection{Semantic Coverage Greedy Selection}
With the proposed optimization objective Eq.~\ref{eq:objective}, dataset pruning can be viewed as selecting a subset of samples whose associated concept sets collectively maximize semantic coverage under a fixed budget. This formulation naturally leads to an iterative greedy optimization strategy based on marginal coverage gain.
At each step of the pruning process, the key question is which sample should be added to the current retained set in order to best improve the overall coverage. Since the objective function rewards the inclusion of previously uncovered concepts, the contribution of a candidate sample depends on the concepts that it introduces beyond those already covered.

Let $\mathcal{S}_t \subseteq \mathcal{D}$ denote the selected subset of samples of the $t$-th step. For any candidate samples $x \in \mathcal{D} \setminus \mathcal{S}_t$ with the corresponding concept graph $G_i=(V_i, E_i)$, its marginal gain with respect to the coverage objective is defined as 
\begin{equation}
\Delta(x_i \mid \mathcal{S}) = F(\mathcal{S} \cup \{x_i\}) - F(\mathcal{S}) = \sum_{v \in V_i \setminus V_{\mathcal{S}} } \phi(v).
\end{equation}
This marginal gain measures the boundary expansion of the current retained subset in the concept space: samples that introduce new, important, and hard-to-cover concepts receive higher scores, while samples whose concepts are already well covered contribute little or no gain. The greedy strategy then selects the sample with the largest marginal gain at each iteration:
\begin{equation}
x^* = \arg \max_{x \in \mathcal{D} \setminus \mathcal{S}_t} \Delta(x \mid \mathcal{S}_t),
\end{equation}
which will be added to the selected subset $\mathcal{S}_{t+1} \leftarrow \mathcal{S}_t \cup \{x^*\}$ for the next iteration. This procedure is repeated until the target budget is reached. Please refer to Algorithm~\ref{alg:mcl} for the whole process.

The necessity of a greedy selection strategy stems directly from the nature of our coverage-based objective. Traditional pruning methods typically assign a static, independent importance score to each sample, which does not consider the selected subset as a whole. This scoring manner inevitably leads to select clusters of redundant samples that share the same high-value concepts. 
In contrast, our objective $F$ inherently evaluate the overall concept coverage of the selected subset.
However, Eq.~\ref{eq:objective} is typically an NP-hard optimization problem, and it is unacceptable to compute the set-dependent $F(\mathcal{S})$ for $b\choose {|\mathcal{D}|} $ possible choice.
To solve Eq.~\ref{eq:objective} more efficiently, we resort to such an iterative greedy algorithm.
Our greedy strategy evaluates the dynamic marginal gain of each candidate, specifically its ability to introduce previously uncovered concepts. Because the value of a sample is explicitly conditioned on the  state of the currently retained subset, the greedy rule continuously targets the data semantic frontier. 
\begin{figure}[t] %
    \centering
    \includegraphics[width=0.9\textwidth]{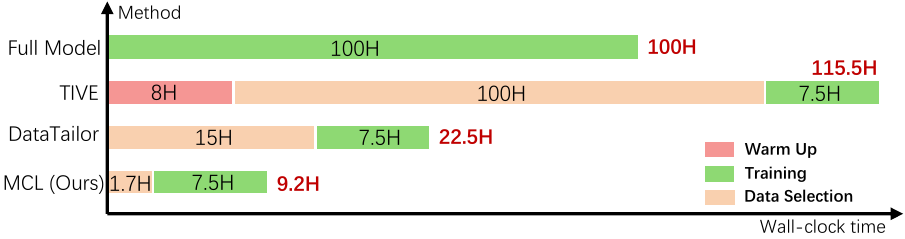}
    \captionsetup{width=0.9\textwidth}
    \caption{Wall-clock time comparison with recent data pruning methods~\cite{tive, datatailor}. Results are collected on a device with four NVIDIA RTX 3090 GPUs. }
    \label{fig:wrap-right}
\end{figure}
\noindent
This dynamic evaluation ensures that every newly selected sample maximally complements, rather than repeats, the existing concepts, which is the key to  eliminating redundancy in large, repetitive datasets.
The greedy selection rule depends only on set differences and concept importance scores, allowing efficient incremental updates. 
In practice, we further accelerate the pruning process by partitioning the dataset into chunks and evaluating marginal gains in parallel across candidate samples, making the approach efficient and scalable to web-scale datasets. For examples, on LLaVA-1.5-mix-665k~\cite{liu2024improved} dataset, we partition the whole dataset into eight chunks and conduct greedy data pruning in parallel. As reported in the Fig.~\ref{fig:wrap-right}, the proposed MCL only introduce a 1.7-hour overhead while the SOTA methods like TIVE~\cite{tive} spend days to conduct data pruning. 
The reported overheads include both the graph building and the selection time.

\begin{algorithm}[t]
\caption{Greedy Concept Coverage Pruning (MCL)}
\label{alg:mcl}
\begin{algorithmic}[1]
    \Statex \textbf{Input:} Dataset $\mathcal{D}=\{x_i\}^N_{i=1}$, concept graph set $\{G_i=(V_i, E_i)\}$ for each $x_i \in \mathcal{D}$, dataset-level concept graph $G_{\mathcal{D}}$, concept node importance $\phi(v)$ and target budget $b=(1-p)N$.
    \Statex \textbf{Output:} Retained sample subset $\mathcal{S}$.
    \smallskip
    \State $\mathcal{S}_0 \gets \emptyset$, $V_{\mathcal{S}_0} \gets \emptyset$. \Comment{Initialize}
    \smallskip
    \For{$t$ \textbf{in} $[0, b]$} 
        \smallskip
        \For{each $x_i \in \mathcal{D} \setminus \mathcal{S}_t$}
            \State Get $\phi(v)$ based on $G_{\mathcal{D}}$ for all nodes in $V_i$. 
            \State Get the marginal gain $\Delta(x_i \mid \mathcal{S}_t) \gets \sum_{v \in V_i \setminus V_{\mathcal{S}_t} } \phi(v)$.
        \EndFor
        \smallskip
        \State Pick the optimal sample $x^* \gets \operatorname*{argmax}_{x} \Delta(x \mid \mathcal{S}_t)$.
        \State $\mathcal{S}_{t+1} \gets \mathcal{S}_t \cup \{x^*\}$ . \Comment{Update retained subset}
        \State $V_{\mathcal{S}_{t+1}} \gets V_{\mathcal{S}_t} \cup V_{x^*}$.
    \EndFor
    \smallskip
    \State \Return $ \mathcal{S} \leftarrow \mathcal{S}_b$
\end{algorithmic}
\end{algorithm}

\section{Experiment}

\subsection{Datasets \& Backbone}
To comprehensively evaluate the effectiveness and generalization of our dataset pruning framework, we conduct experiments on both multimodal vision–language instruction tuning~(VIT) and object detection, a classic pure vision task. This dual setting allows us to assess whether our method can generalize to fundamentally different learning paradigms for multimodal and only image data.

\smallskip
\noindent \textbf{Vision–Language Instruction Tuning.} For multimodal instruction tuning, we adopt the LLaVA-1.5-mix-665k dataset~\cite{llava}. This dataset contains approximately 665K instruction-following samples collected from 12 vision–language tasks, covering a wide range of visual reasoning and perception abilities. Its scale, diversity, and heavy semantic redundancy make it a representative benchmark for evaluating dataset pruning strategies in large-scale multimodal training. We employ the LLaVA-v1.5-7B~\cite{llava} model as our backbone.

\smallskip
\noindent \textbf{Object Detection.}
To further examine the robustness and task-agnostic nature of our approach, we additionally evaluate MCL on the COCO 2017 object detection dataset~\cite{coco}. COCO 2017 contains 118k training images and 5k validation images annotated with 80 object categories. 
There are 7 instances per image on average, ranging from small to large on the same images.
We conduct all experiments using DETR~\cite{detr} with a ResNet-50~\cite{resnet} backbone.
Compared to single-label image classification dataset such as ImageNet~\cite{imagenet}, object detection requires both category recognition and precise spatial localization, and thus relies on richer and more fine-grained visual semantics. This makes COCO more suitable for assessing whether preserving semantic coverage can effectively support complex visual task.
Although object detection is a purely vision-based task, COCO 2017 provides an additional advantage: each training image is paired with five human-written captions. These captions offer a compact, high-level semantic abstraction of visual content, enabling efficient concept extraction. 

Together with the VIT setting, experiments on COCO 2017 demonstrate that our coverage-based pruning strategy generalizes beyond multimodal instruction learning and remains effective in a classical vision task with substantially different supervision and learning paradigm.

\noindent \textbf{Evaluation Benchmarks.}
For VIT, we evaluate the pruned methods on a set of representative benchmarks covering multimodal perception, reasoning, and general visual question answering. Specifically, we include MME~\cite{mme} for perceptual and cognitive evaluation, SEED-Bench-Image~\cite{seed} for comprehensive multimodal assessment across multiple dimensions, POPE~\cite{pope} for hallucination analysis, and MMMU~\cite{mmmu} for challenging scientific reasoning. 
To assess general visual understanding and generalization, we further evaluate on ScienceQA~\cite{scienceqa}, GQA~\cite{gqa}, and TextVQA~\cite{textvqa}, which emphasize reasoning, compositional visual understanding, and text-centric perception, respectively. In all tables and figures, \emph{Rel.} denotes the relative performance to that of the corresponding full-data baseline.
For object detection, we report results on the COCO 2017 validation set using Average Precision (AP).

\vspace{-4pt}
\subsection{Main Results and Discussion}
\vspace{-1pt}

\begin{table}[t!]
  \centering
  \small
    \caption{Comparison to state-of-the-art methods on LLaVA-v1.5-mix-665k dataset with 50k~(7.5\%) and 133k~(20\%) selected data. Our results are shown in the gray block. The best and the second best results are in \textbf{bold} and \underline{underlined}, respectively. We all use the LoRA model due to limited resources. \textsuperscript{$\dagger$} denotes our implementation.}
  \setlength{\tabcolsep}{4pt}
  \resizebox{0.96\linewidth}{!}{
  \begin{tabular}{ l c |c c  c c c | c c c | c }
      \toprule
      \bf Benchmarks & Valid Data & MME-P & MME-C & SEED-I & POPE & MMMU & SQA & GQA & TextQA & Rel. \\
      \midrule
        Baseline & 665k & 1476.9 & 267.9 & 67.4 & 86.4 & 32.8 & 70.0 & 63.0 & 58.2 & 100.0\% \\
        \midrule
        Random & 50k & 1387.5 & 287.5 & 59.7 & \bf 85.7 & 32.2 & 68.4 & 55.0 & 53.1 & 95.7\% \\ 
        Length & 50k & 1357.0 & 265.7 & 47.0 & 82.6 &\underline{33.9} & 60.9 & 55.5 & 45.2 & 89.1\% \\
        EL2N\cite{EL2N} & 50k & 1077.3 & 252.5 & 59.3 & 80.8 & 33.6 & \underline{71.0} & \bf 61.0 & 41.7 & 90.1\%  \\
        GradN\cite{EL2N} & 50k & 1275.4 & 303.6 & 58.3 & 75.7 & 32.4 & 70.9 & 44.9 & 46.0 & 90.5\% \\
        IFD\cite{ifd} & 50k & 1113.4 & 301.8 & 55.1 & 76.7 & 33.0 & 48.2 & 41.9 & 43.6 & 83.6\% \\ 
        InsTag\cite{lu2023instag} & 50k & 1317.1& \bf 345.0& 57.4& 82.1& \bf 34.0& 69.3& 52.5& 53.3& 97.0\% \\
        LESS\cite{xia2024less} & 50k& 1344.8& 281.8& \underline{61.2} & 79.4& 33.0& \underline{ 71.0 } & 53.4& 52.0 & 94.4\% \\
        Self-Filter\textsuperscript{$\dagger$}\cite{selffilter} & 50k & 1331.2 &262.0 &58.2 &78.1 & 31.4 &68.9 &52.8 &50.1 & 91.0\% \\
        TIVE\textsuperscript{$\dagger$}\cite{tive} & 50k &1434.8 &291.5 &\bf 61.6 & \underline{84.9} & 33.3 & \bf 71.2 &56.3 &52.0 & 97.2\% \\
        DataTailor\textsuperscript{$\dagger$}\cite{datatailor} & 50k & \underline{1447.4} & \underline{322.3} & 60.6 & 82.4 & \underline{33.9} & 70.4 & 57.1 & 52.9 & \underline{98.6\%} \\
        \rowcolor[gray]{0.9} 
        MCL(Ours) & 50k & \bf 1455.8 & 318.2 & 60.3 & 84.8 & 33.8 & 70.0 & \bf 57.6 & \bf 53.9 &\bf 99.0\%\\
        \midrule
        COINCIDE\textsuperscript{$\dagger$}\cite{coincide} & 133k & \underline{1496.0} & \underline{298.1} & 62.9 & \underline{86.1} & 32.7 &  69.2 & 59.8 & \underline{55.6} & 99.3\% \\
        ICONS\textsuperscript{$\dagger$}\cite{icons} & 133k & 1487.1 & 295.3 & \underline{63.0} & \bf 87.5 & \underline{33.0} & \bf 70.8 & \bf 60.7 & \underline{55.6} & \underline{99.9\%} \\
        \rowcolor[gray]{0.9} 
        MCL(Ours)\% & 133k & \bf 1501.5 & \bf 308.4 & \bf 63.4 & 85.7 & \bf 33.4 & \underline{70.7} & \underline{60.3} & \bf 56.0 & \bf 100.5\% \\
      \bottomrule
  \end{tabular} }
  \label{tab:llava_75}
\end{table} 

\noindent \textbf{Multimodal Data Selection Results.} 
As shown in Tab.~\ref{tab:llava_75}, experiments on the LLaVA-1.5-mix-665k dataset reveal substantial redundancy in large-scale multimodal instruction data. When retaining only 7.5\% of the original training samples, most pruning methods preserve over 90\% of the full-data performance, and even random selection avoids a severe performance degradation. This observation indicates that VIT training data contain significant semantic redundancy, leading to inefficient training and motivating the need for principled dataset pruning to improve data quality and training efficiency.
Specifically, DataTailor~\cite{datatailor} and our MCL, both of which incorporate similarity-aware selection, consistently outperform approaches. This highlights the importance of preserving the information diversity when pruning large-scale VIT datasets.
More importantly, although DataTailor jointly considers similarity and representativeness, which also adopts the task-adaptive data allocation strategy, our MCL still achieves stronger overall Rel. performance. We attribute the effectiveness of MCL to the fine-grained, disentangled semantic representation induced by our concept graphs, which enables more precise differentiation among semantic factors between samples. As a result, the retained data better supports generalization across various evaluation tasks.
This advantage of our concept graph is further supported by the stability of MCL across benchmarks. Several competing methods exhibit pronounced performance imbalance. For instance, InsTag performs strongly on MME but degrades noticeably on SEED-I and GQA, and LESS excels on SEED-I and ScienceQA but underperforms on MME-C. On the contrary, our MCL maintains near-optimal performance across diverse benchmarks. This provides direct evidence of improved generalization rather than overfitting to specific benchmarks.

In addition to performance, our approach offers significant efficiency advantages. As reported in Fig.~\ref{fig:wrap-right} the entire MCL pruning pipeline including concept graph construction, concept importance estimation, and greedy selection can be completed in 1.7 hours on a 4 * RTX 3090 GPU server. In contrast, methods such as TIVE and DataTailor require expensive pretraining, feature extraction, or clustering procedures. With less than 10\% of the overall time cost, our method retains 7.5\% of the data but still achieves up to 99\% of the full-data performance, demonstrating a substantial improvement in training efficiency.

\begin{table}[t!]
  \centering
    \caption{Comparison to state-of-the-art data pruning methods on COCO 2017 object detection dataset with 70\% selected training data. }
    \vspace{-6pt}
  \setlength{\tabcolsep}{7pt}
  \resizebox{0.8\linewidth}{!}{
  \begin{tabular}{ l | c | c c c c c c c  }
      \toprule
       Method & $AP_{50:95}$ & $AP_{50}$ & $AP_{75}$ & $AP_{small}$ & $AP_{mid}$ & $AP_{large}$  \\
      \midrule
        Baseline & 40.06 & 61.12 & 42.03 & 19.27 & 43.39 & 59.24  \\
        \midrule
        Random & 38.26~$\downarrow$1.80  &	58.91&	39.77&	17.83&	41.19&	56.69  \\
        EL2N & 34.35~$\downarrow$5.71 & 53.29 & 36.14 & 16.58 & 37.64 & 48.54 \\
        InfoBatch & 34.95~$\downarrow$5.11	&53.86	&36.55	&17.00	&38.74	&49.05 \\
        DivBS	&39.74~$\downarrow$0.32	&60.84	&41.98	&18.07	&43.14	&58.78 \\
        PFB &39.69~$\downarrow$0.37	&60.85	&41.74	&18.45	&43.25	&59.07 \\
        \rowcolor[gray]{0.9} 
        Ours &40.12~$\uparrow$0.06	&61.06	&42.27	&18.95	&43.78	&59.13\\
      \bottomrule
  \end{tabular} }
  \vspace{-5pt}
  \label{tab:coco}
\end{table}

\smallskip
\noindent \textbf{Object Detection Results.} 
Beyond multimodal instruction tuning, we further evaluate our approach on the purely vision-based object detection task. As reported in Tab.~\ref{tab:coco}, our method achieves higher performance than the full-data baseline while using 70\% of the COCO 2017 training data, demonstrating that semantic coverage–aware pruning remains effective even when supervision and task objectives differ substantially from instruction tuning on pure vision data.
We also compare our method with general data pruning approaches such as InfoBatch~\cite{infobatch}, DivBS~\cite{DivBS}, and PFB~\cite{pfb}. Among them, DivBS and PFB explicitly promote diversity based on embedding representations. Our consistent performance advantage over these methods suggests that fine-grained, concept-level representations provide a more expressive and effective basis for dataset pruning than coarse, coupled vector embeddings. This result further indicates that the benefits of our semantic coverage formulation extend beyond multimodal data and generalize to classical computer vision tasks.

\begin{figure*}[t!]
  \centering
  \begin{minipage}{0.48\linewidth}
    \centering
    \includegraphics[width=0.95\linewidth]{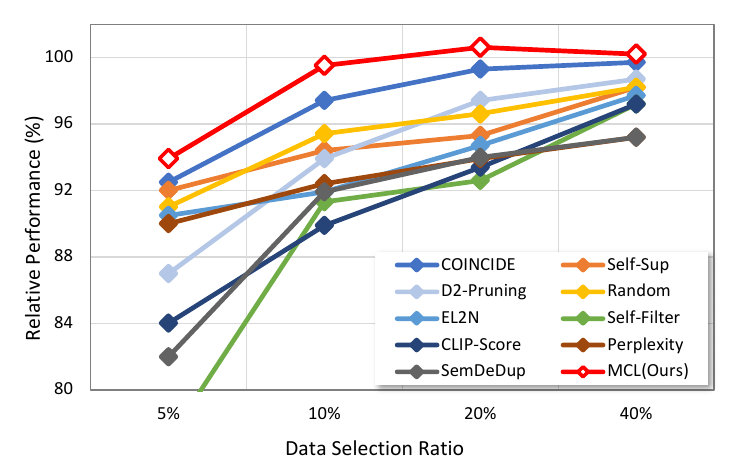} 
    \vspace{-6pt}
    \caption{Relative performance of different data pruning methods at different sampling ratios for the LLaVA-1.5 dataset.  
    }
    \label{fig:whole_ratio}
  \end{minipage}
  \hfill
  \begin{minipage}{0.48\linewidth}
    \centering
    \includegraphics[width=0.98\linewidth]{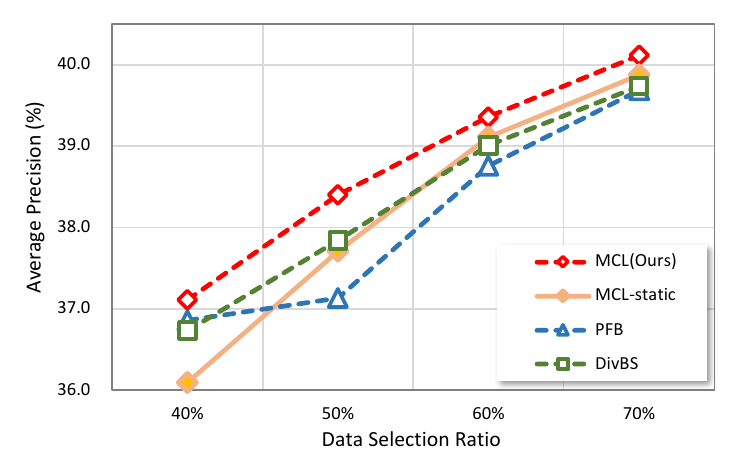}
    \vspace{-6pt}
    \captionof{figure}{ Different selection ratios on COCO 2017 val. set. Static denotes pruning by ranking the sum of node scores.}
    \label{fig:coco}
  \end{minipage}
\end{figure*}

\subsection{Ablation Study}

\noindent \textbf{Performance at Different Selection Ratios.} 
In Fig.~\ref{fig:whole_ratio}, we compare MCL with other recent pruning methods, including COINCIDE~\cite{coincide}, D\textsuperscript{2}-Pruning~\cite{d2}, SemDeDup~\cite{abbas2023semdedup}, and Self-Sup~\cite{sorscher2022beyond}, across a wide range of selection ratios on LLaVA-1.5-mix-665k. MCL consistently outperforms competing approaches under different data budgets, demonstrating the effectiveness of our framework.
In particular, MCL surpasses COINCIDE, which estimates density (rarity) using sample-wise embeddings. Similar trends are observed in Fig.~\ref{fig:coco} on COCO, where MCL based on decoupled concept graphs achieves clear improvements over PFB and DivBS, both of which rely on diversity measured in sample-level embedding space.
These results suggest that concept graphs provide a more reliable representation for measuring semantic diversity than compressed sample embeddings.

\noindent \textbf{Static vs. Greedy Selection.}
We compare our greedy marginal-gain-based pruning with a static alternative that ranks samples solely by their standalone concept importance~(Eq.~\ref{eq:importance_norm}), as shown in Fig.~\ref{fig:coco}. On COCO 2017 object detection, the greedy strategy consistently achieves better performance. Static ranking tends to prioritize samples containing rare concepts because importance scores are computed once which can not evolve along with the selection process. Thus, the final selected subset may overemphasize long-tail concepts while underrepresenting dominant concepts that are also essential for learning. This bias becomes increasingly pronounced as the selection ratio decreases, which leads to a larger performance gap between static ranking and the greedy strategy of MCL. These results indicate that effective pruning requires adaptively balancing long-tail concepts with dominant semantics, which is naturally achieved by the our greedy selection based on marginal gains.

\begin{figure*}[t!]
  \centering

  \begin{minipage}{0.48\linewidth}
    \centering
    \vspace{-2pt}
    \includegraphics[width=0.95\linewidth]{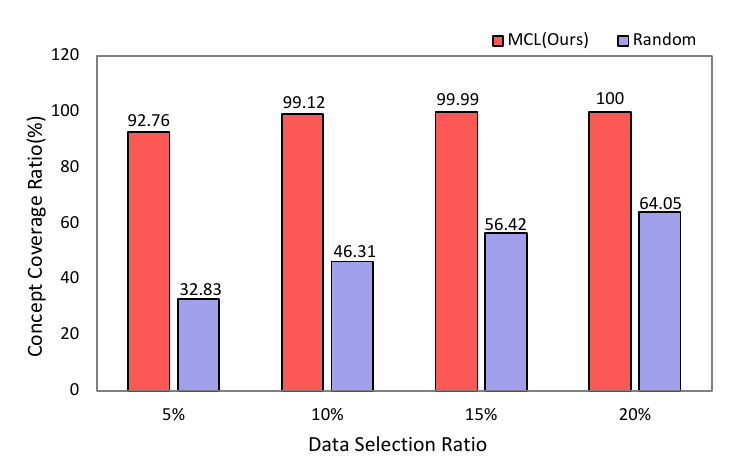}
    \vspace{-6pt}
    \captionof{figure}{Semantic concept coverage ratio~($|V_{\mathcal{S}}|/|V_{\mathcal{D}}|$) of MCL and Random pruning on LLaVA-1.5-mix-665k dataset. }
    \label{fig:coverage}
  \end{minipage}
  \hfill
  \begin{minipage}{0.48\linewidth}
    \centering
    \includegraphics[width=0.95\linewidth]{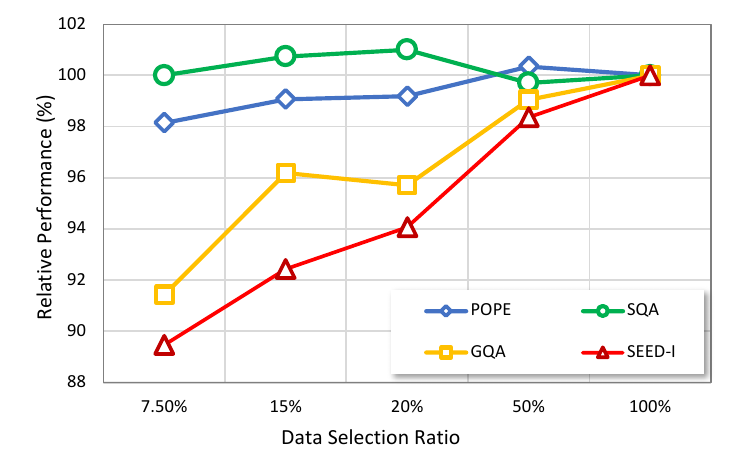} 
    \vspace{-6pt}
    \caption{Relative model performance comparison on different data selection ratios on LLaVA-1.5-mix-665k dataset.
    }   
    \label{fig:bench_ratio}
  \end{minipage}
\end{figure*}

\noindent \textbf{Fine-grained Concept Coverage.}
We investigate whether our method can maintain a high fine-grained semantic coverage at the dataset level. To this end, we compare the concept coverage ratio between the subset selected by MCL and a randomly pruned subset. Quantitative results in Fig.~\ref{fig:coverage} show that MCL covers a substantially larger portion of semantic concepts, confirming that coverage-aware pruning remains effective beyond simple data reduction.

\begin{figure}[t]  %
    \centering
    \tiny
    \begin{tabular}{cc}
    \includegraphics[height=0.42\textwidth]{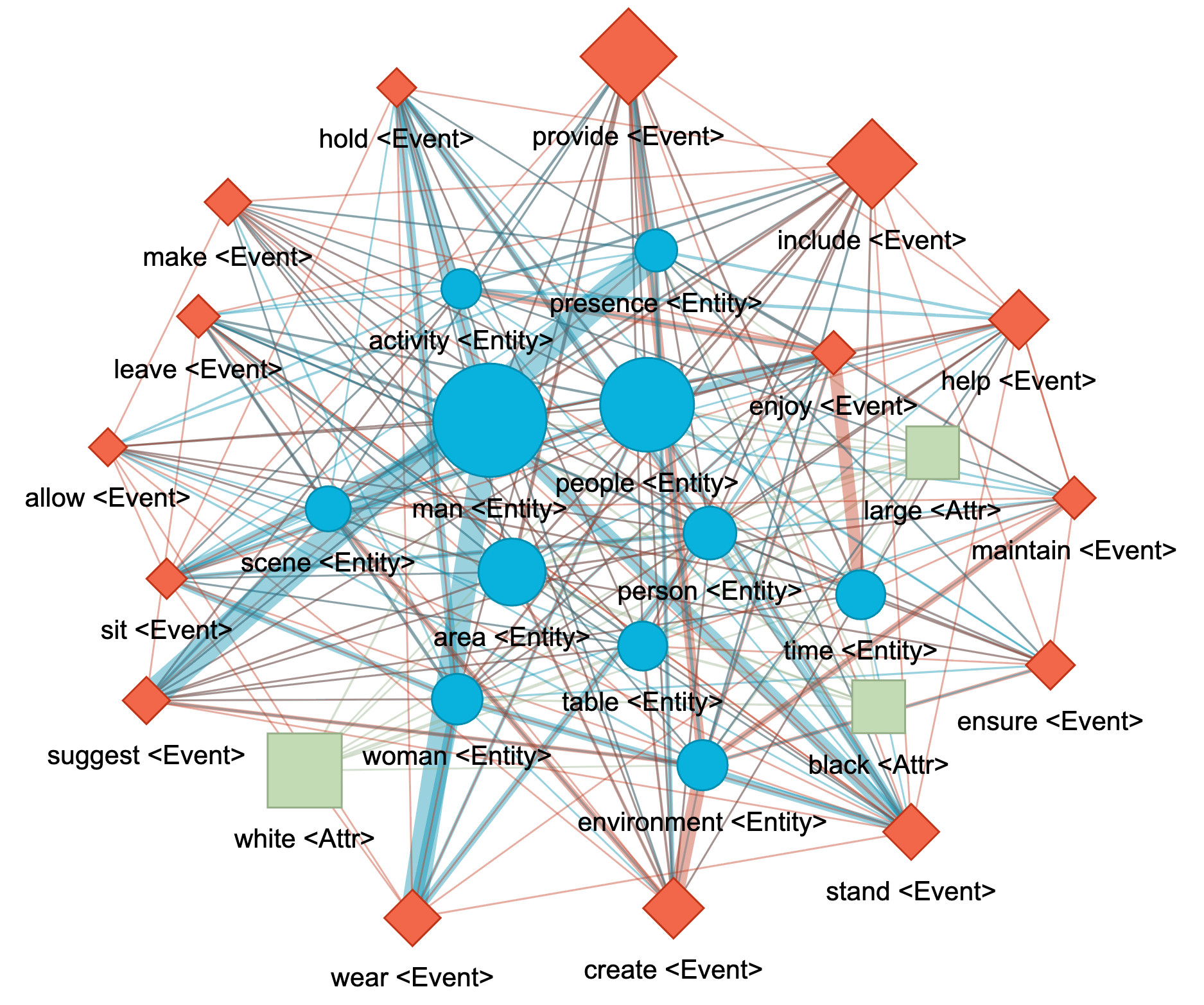} 
    & 
    \includegraphics[height=0.42\textwidth]{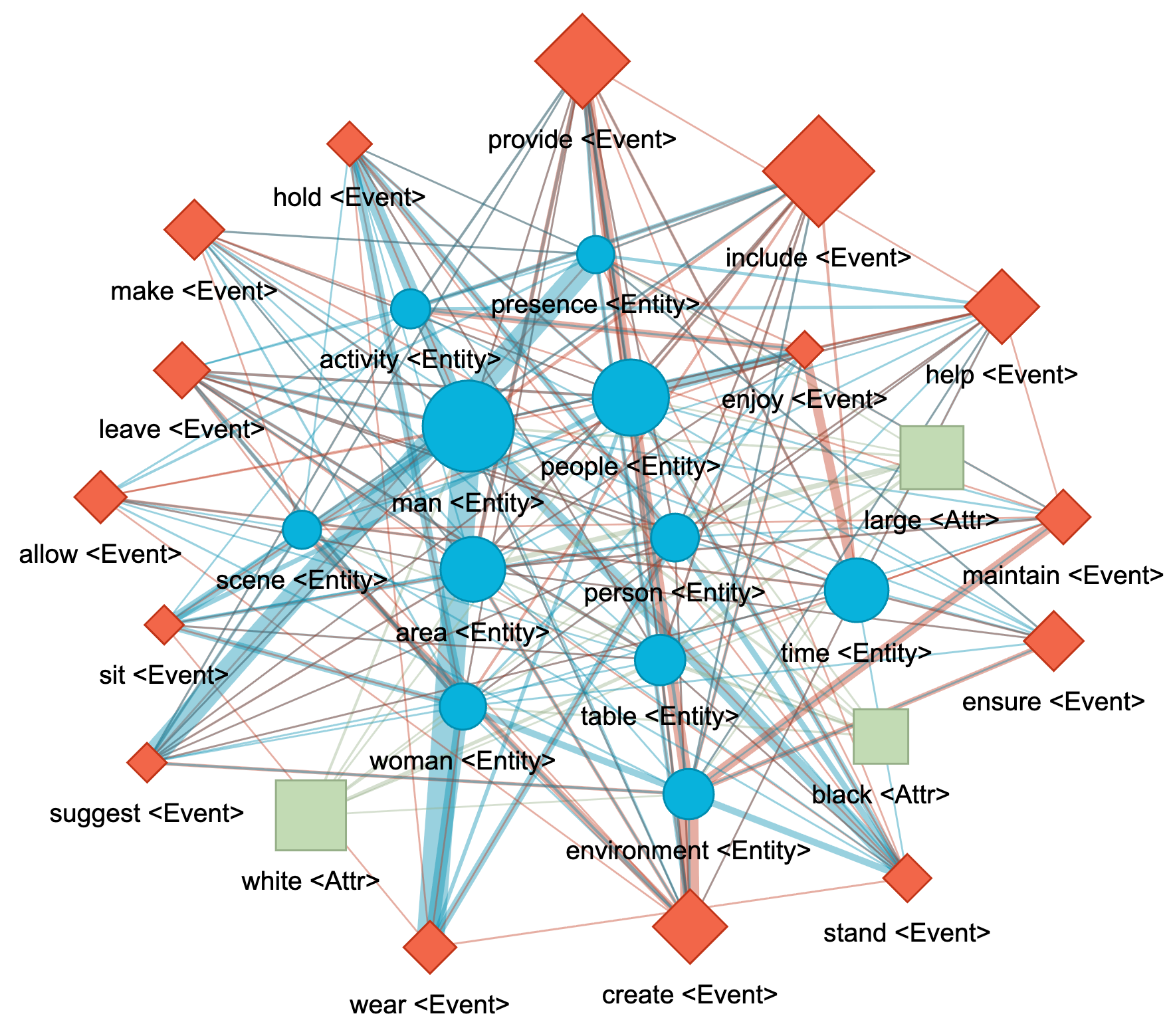} 
    \\
    (a) Concept graph before pruning & (b) Concept graph after pruning
    \end{tabular} 
    \caption{ Visualization of the dataset graph (a) before and (b) after pruning. The size of each node is proportional to the fraction of samples containing the concept. Concept distribution becomes more balanced after pruning.}
    \label{fig:vis}
\end{figure}

\noindent \textbf{Concept Visualization.}
To further provide an intuitive understanding of how 
semantic distribution changes after pruning, we visualize the dataset concept graph before and after pruning. In the Fig.~\ref{fig:vis}, we distinguish the three 
types of concept nodes using different colors and shapes. Node sizes are proportional to their relative fraction across the dataset, reflecting how often each concept appears in the retained samples. For clarity, we focus on the top 30 most frequent concept nodes in the original dataset graph and the edges among them.
Comparing the concept graphs before and after pruning reveals a clear redistribution of the  semantic concepts. 
Highly redundant concepts, such as frequent entity nodes (e.g., \textit{man, people}), occupy a smaller proportion after pruning, while less frequent but semantically informative event (e.g., \textit{leave, create}) and attribute nodes gain increased relative prominence. This shift indicates that our method effectively suppresses over-represented concepts while amplifying those under-represented ones, leading to a more balanced semantic distribution.
Importantly, this rebalancing is not achieved by discarding dominant concepts entirely, but by moderating their prevalence to make room for rarer concepts. Consequently, the retained subset maintains broad semantic coverage while avoiding excessive redundancy. This observation aligns with our coverage-based objective and provides direct evidence that concept-level pruning can reshape dataset distributions in a principled and interpretable manner.

\smallskip
\noindent \textbf{Effect of Pruning Ratios on Different Benchmarks.}
Fig.~\ref{fig:bench_ratio} shows that benchmarks exhibit markedly different sensitivities to the data selection ratio. For POPE and ScienceQA, performance saturates rapidly: with only 7.5\% of the data, MCL already reaches or even slightly exceeds the full-data baseline. This indicates that the training set contains substantial redundant data corresponding to these tasks, where a small subset can cover the dominant semantic patterns required for the benchmarks.
In contrast, SEED-I and GQA benefit from increased data ratios, reflecting they are more difficult and requires a broader concept distribution. Notably, performance does not increase monotonically with more data on all benchmarks, suggesting that coverage-aware selection can be more effective than retaining larger but redundant subsets.

\section{Conclusion}
We present MCL, a coverage-aware dataset pruning framework that models large-scale multimodal data through interpretable, fine-grained concept graphs. By decoupling samples into entity, event, and attribute concepts and performing greedy selection based on set-dependent marginal gains, MCL effectively reduces redundancy while preserving semantic diversity. Extensive experiments on both multimodal instruction tuning and object detection demonstrate that MCL achieves strong performance under aggressive pruning, offering substantial improvements in data efficiency with minimal computational overhead.

\bibliographystyle{splncs04}
\bibliography{main}

\end{document}